\documentclass[runningheads,a4paper]{llncs}

\usepackage{amssymb}
\usepackage{graphicx}
\usepackage{caption}
\usepackage{floatrow}
\usepackage{subfigure}

\usepackage{url}
\urldef{\mailsa}\path|dishriva@in.ibm.com, schaudhury@gmail.com, jayadeva@ee.iitd.ac.in |
\newcommand{\keywords}[1]{\par\addvspace\baselineskip
\noindent\keywordname\enspace\ignorespaces#1}

\begin{document}

\mainmatter  % start of an individual contribution

% first the title is needed
\title{A Data and Model-Parallel, Distributed and Scalable Framework for Training of Deep Networks in Apache Spark}

% the name(s) of the author(s) follow(s) next
%
% NB: Chinese authors should write their first names(s) in front of
% their surnames. This ensures that the names appear correctly in
% the running heads and the author index.
%
\author{Disha Shrivastava\thanks{This work was done when author was student at IIT Delhi}\inst{1} \and Santanu Chaudhury\inst{2,3} \and Dr. Jayadeva\inst{3}}
\institute{IBM Research India\\
\and CSIR-CEERI\\
\and IIT Delhi\\
\mailsa,}

\maketitle

\begin{abstract}
Training deep networks is expensive and time-consuming with the training period increasing with data size and growth in model parameters. In this paper, we provide a framework for distributed training of deep networks over a cluster of CPUs in Apache Spark. The framework implements both Data Parallelism and Model Parallelism making it suitable to use for deep networks which require huge training data and model parameters which are too big to fit into the memory of a single machine. It can be scaled easily over a cluster of cheap commodity hardware to attain significant speedup and obtain better results making it quite economical as compared to farm of GPUs and supercomputers. We have proposed a new algorithm for training of deep networks for the case when the network is partitioned across the machines (Model Parallelism) along with detailed cost analysis and proof of convergence of the same. We have developed implementations for Fully-Connected Feedforward Network, CNN, RNN and LSTM architectures. We present the results of extensive simulations demonstrating the speedup and accuracy obtained by our framework for different sizes of the data and model parameters with variation in the number of worker cores/partitions; thereby showing that our proposed framework can achieve significant speedup (upto 11X for CNN) and is also quite scalable.

\end{abstract}

\section{Introduction}
Many of the recent advances in deep networks involve fitting large models to massive datasets to obtain good results. Training these networks is time-consuming, the training period ranging from days to even weeks. Thus distributed implementation of deep networks in a way that the data can be partitioned and stored in multiple machines (Data Parallelism) becomes important. Additionally, the model parameters need to be distributed across multiple machines (Model Parallelism). This calls for an efficient mechanism for update of model parameters in a distributed setting reducing the communication overhead, yet maintaining the accuracies typical in non-distributed environments.

Much work has been done to build distributed frameworks for training deep networks. DistBelief from Google\cite{dean2012large} and Project Adams \cite{chilimbi2014project} from Microsoft are both distributed frameworks meant for training large scale models for deep networks over thousands of machines and utilizing both data and model parallelism. Training deep networks by a model-parallel system on a GPU cluster using MPI over Infiniband has been proposed in \cite{catanzaro2013deep}. A central parameter server model has been used in \cite{li2014scaling}\cite{ho2013more} in which the worker nodes asynchronously fetch and update the gradients from a master node which holds the updated model parameters; though their systems are better suited for sparse gradient updates. Google's recent  TensorFlow \cite{abadi2016tensorflow} framework is an improved version of DistBelief system which can be used for distributed training of deep networks by specifying them as computational flow graphs. FireCaffe \cite{iandola2015firecaffe} uses a cluster of GPUs to scale deep networks over Infiniband/Cray interconnects. The systems described above show high performance in terms of speedup and accuracy and a fine grained control over scheduling. However, due to their demanding communication requirements, they are unlikely to exhibit the same scaling on a cluster of CPUs. Moreover, these systems are highly customized in terms of either requirement of particular hardware or software tools and hence they are difficult to integrate with general-purpose batch computational frameworks such as Spark and MapReduce. The use of GPUs is restricted because they are expensive and with large model parameters the training speedup is small as the model does not fit into the GPU memory and CPU-GPU transfers prove to be a significant bottleneck. Recently, \cite{kim2016deepspark} and \cite{moritz2015sparknet} integrated Spark with Caffe. Their model however implemented just data parallelism without model parallelism.

In this work, we produce a distributed and scalable framework for training of deep networks implementing both data and model parallelism using only Apache Spark \cite{spark}. The reason for choosing Spark is because it provides an open-source, generic batch computational framework which makes it easy to implement distributed optimization algorithms over a cluster of commodity hardware. Spark's MLlib/ML libraries \cite{sparkmllib} have officially released implementations only for Multi-Layer Perceptron classifiers (Fully Connected Neural Nets) that too just implementing data parallelism. They do not have implementations for CNN, RNN and LSTM; which makes our implementation more unique and important. To our knowledge, this is the only work which implements both data and model parallelism in Apache Spark for any of the network architectures. Moreover, we have obtained significant speedup (upto 11X in case of CNN) using our scalable, distributed framework in Spark. The main contributions of this work can be summarized as below:
\begin{itemize}
  \item Proposal of a new algorithm for training of deep networks for the case when the network is partitioned across the machines (Model Parallelism); along with detailed cost analysis and mathematical and experimental proof of convergence of the algorithm.
  
  \item Implementation of a generic framework which can be used for training Fully-Connected Feedforward Networks, CNN, RNN and LSTM architectures.
  
  \item Thorough experimental analysis with variations in data and model sizes and number of worker cores/nodes/partitions, important for understanding the scalability and distributed performance of our framework.
\end{itemize}

\section{Methodology}

\subsection{Downpour SGD}
 Downpour SGD was proposed by \cite{dean2012large} which provides asynchronous and distributed implementation of stochastic gradient descent.
The training instances are divided across different machines \textbf{(Data Parallelism)}. Each machine has its own copy of the whole model/neural network called model replica. If the network is big enough, it can be divided across different machines \textbf{(Model Parallelism)}. Each machine operates on its own set of  data and model replica to compute changes in parameters (weights and biases). These values are periodically sent to a central parameter server where the parameter update is done according to the learning rule. The updated parameters are fetched from the parameter server asynchronously with some delay by the individual machines.

The gradient updates may be out-of-order and the weights may be out of date. However, it has been shown that the delay may be harmful initially but is harmless as the number of iterations and the size of training data increases \cite{zinkevich2010parallelized,zinkevich2009slow}. The use case for data parallelism is when we have massive amount of training examples and the use case for model parallelism is when the neural network is too big to fit into the memory of a single machine. Hence, this training model allows us to work with deep networks with large data and model sizes efficiently and achieve significant speedups.

\begin{figure*}[t]
%\centering
\subfigure[Forward Pass]{\includegraphics[width=0.3\textwidth]{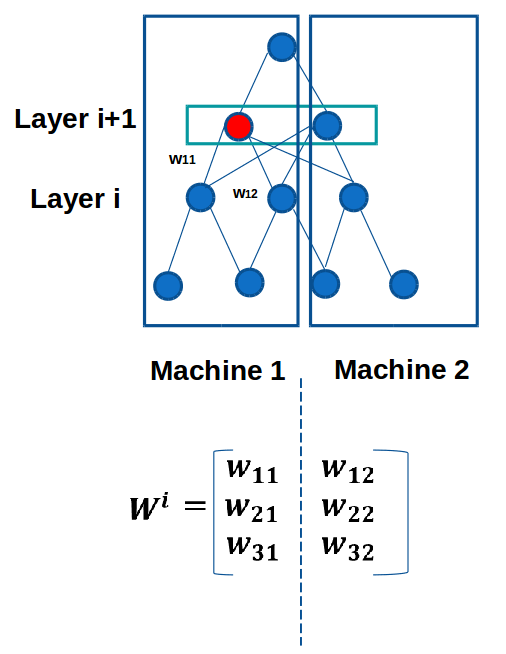}\label{fig:f1}}
\subfigure[Backward Pass]{\includegraphics[width=0.3\textwidth]{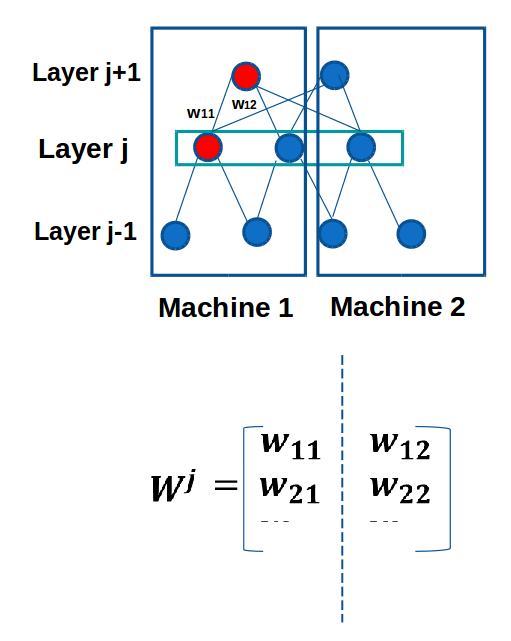}\label{fig:f2}}
\caption{Illustration of distributed backpropagation in case of network partitioning across the machines}%\label{fig:models}
\end{figure*}

\subsection{Our Algorithm for Model Parallelism}\label{sec}
Though \cite{dean2012large} put forth the idea of Model Parallelism, they never discussed about the exact algorithm used for training the network. In this work, we propose and implement a simple algorithm for distributed backpropagation for the case when the deep network is too big to fit into the memory of a single node and hence the network is divided across different nodes. The term processes used below represent computational units (threads/processes) and are distributed across the nodes, i.e., a single node may contain more than one process. One of the
processes is designated as master and the rest as slaves. The two phases of the distributed backpropagation can be described as below:
\begin{itemize}
    \item \textbf{Forward Pass:} The weight matrix is partitioned vertically in a uniform way(no.of columns/no. of processes) and distributed across the machines as shown in Fig. \ref{fig:f1}. The main steps in the algorithm at layer i+1 can be described as below:
\begin{itemize}
    
    \item[1.] The master process broadcasts layer i output vector, $a^i = [a_1  a_2  a_3]$ to all other processes
    \item[2.] Each process computes its subset of layer i+1 output vectors using the weights it has; e.g. On Machine 1 in Figure \ref{fig:f1} for the neuron highlighted in red, the output is calculated as

 \begin{equation}
    a_1^{i+1} = \sigma([a_1  a_2  a_3]\odot [w_{11}  w_{21}  w_{31}])
 \end{equation}  
    \item[3.] The master gathers from all the processes layer i+1's output vectors $a^{i+1}$
    \item[4.] We repeat from Step 1 till the output layer is reached.
\end{itemize}

\item \textbf{Backward Pass:} The algorithm at layer j can be written in the below steps:
\begin{itemize}
\item[1.] Master process broadcasts the error vector of the preceding layer, $\delta^{j+1}$ to all processes.
\item[2.] Each process computes the weight changes for its subset using the output vectors obtained during the forward pass; e.g. in Fig. \ref{fig:f2} the change in weight $w_{11}$ is calculated as 

\begin{equation}
\partial{E}/ \partial{w_{11}} = \delta_1^{j+1} * a_1^j
\end{equation} 
\item[3.] Each process computes its contribution to the error vector at layer j (Here, $\sigma'(in^j)$ is the gradient of the sigmoid activation function at layer j and $W^j$ represents the weight matrix between layer j and j+1) 

\begin{equation}
\delta^j = \sum_{j+1} \delta^{j+1}*W^j*\sigma'(in^j)
\end{equation} 
\item[4.] Master sums up the individual contribution of $\delta^j$ from each process
\item[5.] We repeat from Step 1 till the input layer is reached.
\end{itemize}
\end{itemize}

\subsection{Cost Analysis of the Model}
To do a theoretical cost analysis of our model, we make use of the method given in \cite{pethick2003parallelization} which is reproduced here for convenience of understanding. There are two factors which influence the execution time of our model:the time taken for computation ($T_{comp}$) and the time taken for communication ($T_{comm}$). Ignoring the computation time, the general equation for the cost analysis is:

\begin{equation}\label{eq6}
    T_{comm} = K[T_{lat} + N*T_{data}] ,
\end{equation} 

where $T_{lat}$ is the latency associated with sending a message, $T_{data}$ is the time required to transmit a unit of data which is inversely proportional to the network speed, K is the total number of messages sent per epoch, and N is the number of units of data sent. Let us assume the number of processes to be F and the number of training examples to be M. For each of the training examples, initially the master process sends the entire input data and the processes' portion of the output labels to rest of the slave processes generating (F - 1) messages (Eq. \ref{eq1}). In these equations, $b_i$= number of neurons at layer i, $\delta^i$ = error vector at layer i, n = total number of layers in the network and S(x) = size of the data structure x. During the forward pass, for each of the hidden layers, each process sends the output of its neurons to all other processes. Assuming this communication to be an all-to-all broadcast in a hypercube structure, each single process F requires $\log_2 F$ messages to be sent generating a total of $F \log_2 F$ messages (Eq. \ref{eq2}). The backward pass results in generation of two types of messages per hidden layer. The master process sends the error vector from previous layer $\delta^{j+1}$ to all other processes which is followed by each process sending its portion of the error vector  of the current layer $\delta^{j}$ to the master resulting in 2(F - 1) messages (Eq. \ref{eq3}). Therefore, the value of K is calculated to be:

\begin{equation}\label{eq5}
    K=M[(F-1) + \sum_{i=1}^{n-2} ( F \log_2 F + 2(F-1))
\end{equation} 

    \begin{itemize}
        \item Number of units of data sent during initialization: 

      \begin{equation}\label{eq1}
    N_1 = (F-1)[ S(b_0) + S(\frac{b_{n-1}}{F})]
    \end{equation}  
     \item Number of units of data sent during Forward Pass: 

      \begin{equation}\label{eq2}
    N_2 =  \sum_{i=1}^{n-2} F \log_2 F * S(\frac{b_i}{F})
    \end{equation} 
     \item Number of units of data sent during Backward Pass: 

      \begin{equation}\label{eq3}
    N_3 =  \sum_{i=1}^{n-2} (F-1)*S(\delta^{i+1}) + (F-1) * S(\frac{\delta^i}{F})
    \end{equation} 
    \item Total Number of units of data: %$N = \frac{N_1 + N_2 + N_3}{K}$

      \begin{equation}\label{eq4}
    N = \frac{N_1 + N_2 + N_3}{K}
    \end{equation} 
        
    \end{itemize}
    
Putting the values from Eq.(\ref{eq5}),(\ref{eq1}),(\ref{eq2}),(\ref{eq3}) and (\ref{eq4}) in Eq. (\ref{eq6}), we get the theoretical cost of our model. We can see that with increase in the number of nodes (or F), there is a trade off between the degree of parallelization which we can achieve and the communication overhead.

\subsection{Proof of Convergence of the Algorithm}
We derive an upper bound on the expected regret and hence prove that our model of delayed stochastic gradient descent for the case when both data and network is distributed across different machines converges. The proof is based along the lines discussed in \cite{zinkevich2009slow} and the details of it can be found in the Appendix.

\section{Experiments}
\textbf{Cluster Setup}:
In order to carry out our experiments, we had setup a cluster of five nodes in our lab. Out of these five machines, four machines had Intel Xeon E5-2600 processor/64GB RAM/20 cores, and the remaining one has the same processor but 32GB RAM/8 cores; thereby giving us a total of 88 cores. One of them is designated the master node and the rest as slaves/workers. We have formed a multi-node hadoop cluster for the Distributed File System Storage(HDFS) part and Apache Spark in distributed mode has been setup on top of that.

\textbf{Implementation Details}: 
%\subsection{Implementation Details}
In the first step, training and test data was stored in the form of RDD \cite{zaharia2012resilient}. We distributed the training data RDDs to different machines and gradients were calculated independently on individual machines. The model if big enough was distributed across different machines by partitioning the weight matrix column-wise and training by the proposed algorithm for distributed backpropagation. The gradients were then collected from different machines at the master node and update in parameters was done according to the learning rule. This process was carried on for repeated number of iterations till the model converges. The prediction was made on a separate test set. For different types of neural networks changes are made just in the way forward and backward passes are performed in individual machines and some hyperparameters are changed accordingly. We use threads in individual machines to attain more speedup. We made use of JBlas \cite{jblas} and Mallet \cite{mallet} libraries for linear algebra and optimization, respectively. In case of network partitioning, communication between processes in the same machine is via threads and for processes in different machines via Akka framework \cite{thurau2012akka} called from within Spark jobs. The results obtained were averaged over multiple runs due to variations in CPU loads and network latencies. All the results were obtained using only CPUs (no GPUs).

The extended version of MNIST dataset \cite{mnist} was used as a benchmark dataset for carrying out experiments on Fully Connected Nets and CNN. It contains about 8.1 million samples with 28X28 binary images of handwritten digits from 0 to 9 (10 output classes). For the task of character level prediction using RNN and LSTM, we used the Project Gutenberg \cite{hart1971project} dataset which contains the complete works of Shakespeare.

\textbf{Network-specific Training and Architecture Details}: 
We experimented with different number of hidden layers, number of neurons in each layer and kernel sizes for each of the networks. We state below the specific details which were used to obtain the results mentioned in the next section. We used a three layer Fully Connected net with 784 (28 X 28) neurons at the input layer, 480 and 160 neurons in the first and second hidden layers respectively, followed by 10 neurons at the output layer corresponding to the 10 output classes. For the model parallelism part, we varied the number of neurons and number of layers to generate suitable number of model parameters. Among all the gradient descent techniques, Conjugate Gradient Descent was found to be the fastest and showed the best results. In CNN, the three convolution layers contained kernel of size 5 X 5 with 6 feature maps, kernel of size 5 X 5 with 12 feature maps and kernel of size 4 X 4 with 12 feature maps respectively; with the first and second convolution layers followed by a mean pooling layer of size 2 X 2. The last layer was a FC layer with softmax. Mini-batch gradient descent(with momentum) with batch size of 16 was used. For RNN and LSTM, we used a randomly sampled sequence of 1000 characters as input for our network. The number of neurons at the input and output layer of the network equals the size of the vocabulary (number of valid characters). The number of LSTM/RNN units at first and second hidden layers was kept as 200 and 100, respectively. The output was obtained by sampling 300 characters from the network at the end of each epoch. The LSTM block implemented was the one discussed in \cite{graves2012neural}, consisting of four gates and no peephole connections. Mini-batch gradient descent with batch size of 32 (tBPTT algorithm for training) along with RMSprop was used for optimization. Our developed implementation was a multilayer RNN/LSTM which is capable of handing variable length input and output sequences by padding and masking.
    
\section{Results and Analysis}
\begin{figure*}[t]
%\centering
\subfigure[Error Convergence Graph]{\includegraphics[width=0.385\textwidth]{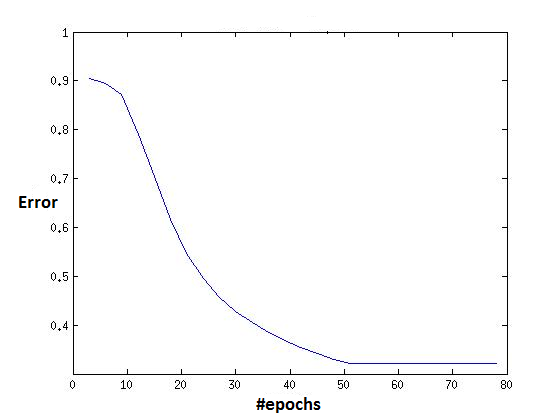}\label{fig:f5}}
\subfigure[Gradient value convergence]{\includegraphics[width=0.385\textwidth]{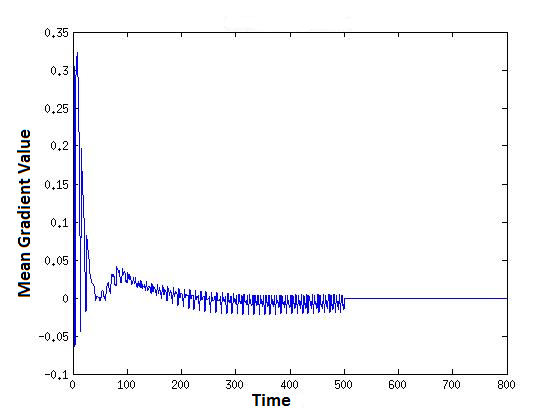}\label{fig:f6}}
\caption{Plots demonstrating the correctness of the algorithm}%\label{fig:models}
\end{figure*}

\begin{figure*}[t]
  \subfigure[Time Performance]{\includegraphics[width=0.455\textwidth]{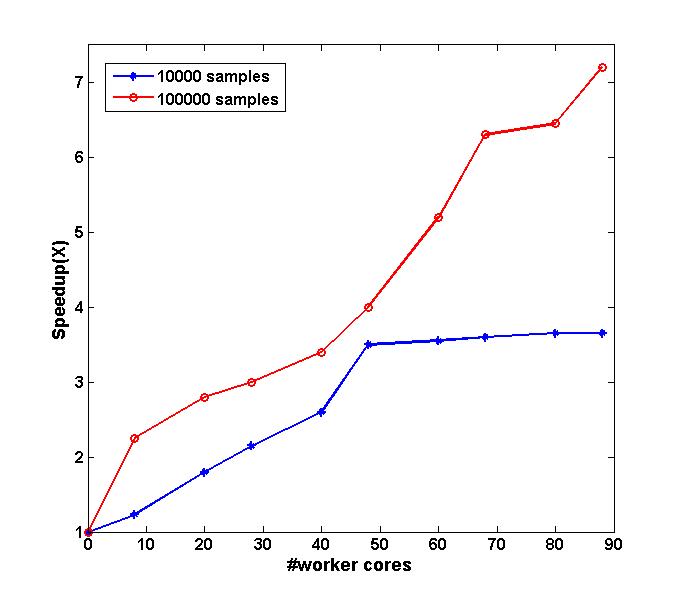}\label{fig:f36}}
 \subfigure[Accuracy Performance]{\includegraphics[width=0.455\textwidth]{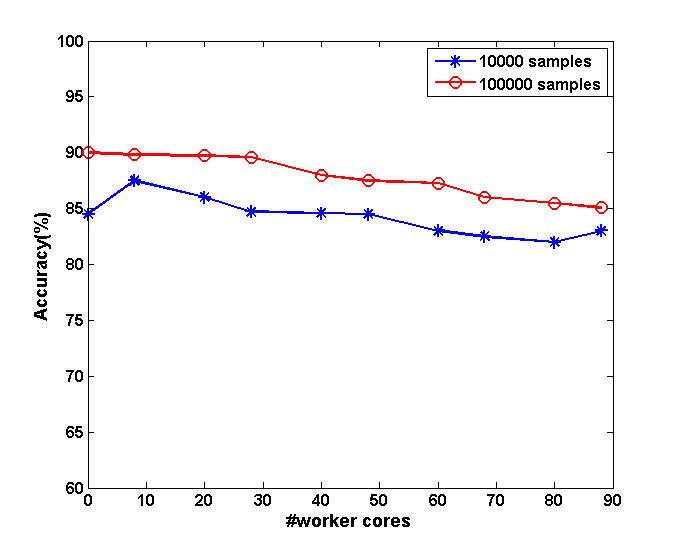}\label{fig:h}}
  \caption{Performance of the cluster for FC Nets [48 cores=20+20+8=3 nodes]}
\end{figure*}

\begin{figure*}[h!]
  \subfigure[Time Performance]{\includegraphics[width=0.465\textwidth]{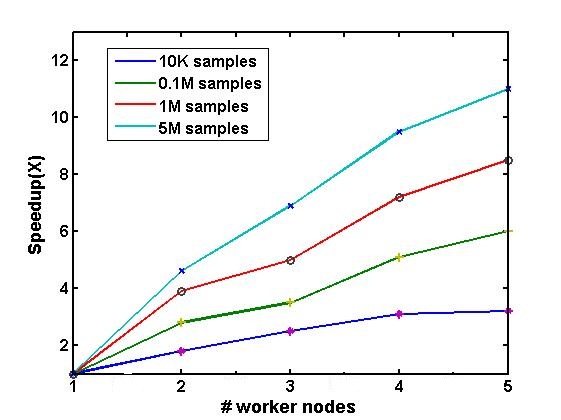}\label{fig:f33}}
  \subfigure[Accuracy Performance]{\includegraphics[width=0.465\textwidth]{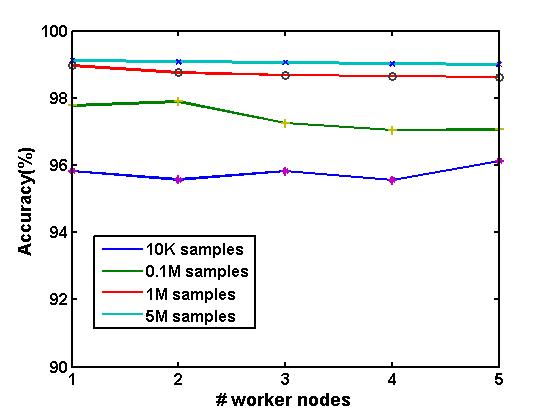}\label{fig:f34}}
  \caption{Performance of the cluster for CNN}
\end{figure*}

\begin{figure}[htb!]
\ffigbox[\textwidth]
  {
   \begin{floatrow}
    \ffigbox[\linewidth]
      {\captionof{figure}{Performance for different sizes of model parameters}\label{fig:h8}}
      {\includegraphics[width=1.0\linewidth]{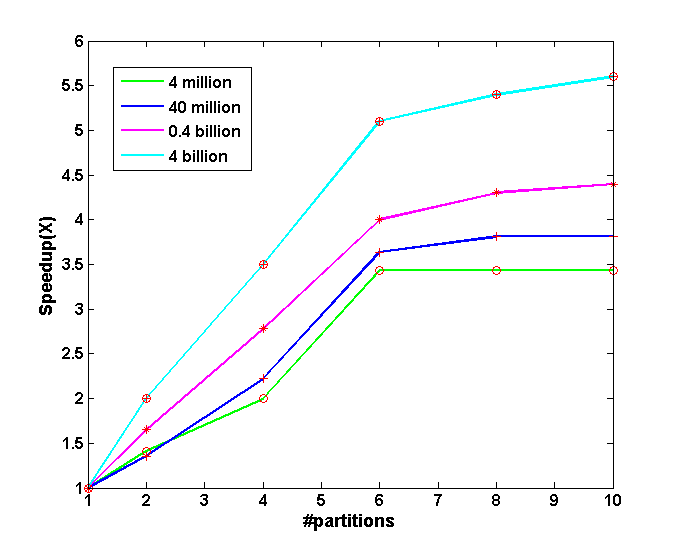}}
    \ffigbox[\linewidth]
      {\captionof{figure}{Scalability Performance for LSTM}\label{fig:h9}}
      {\includegraphics[width=0.9\linewidth]{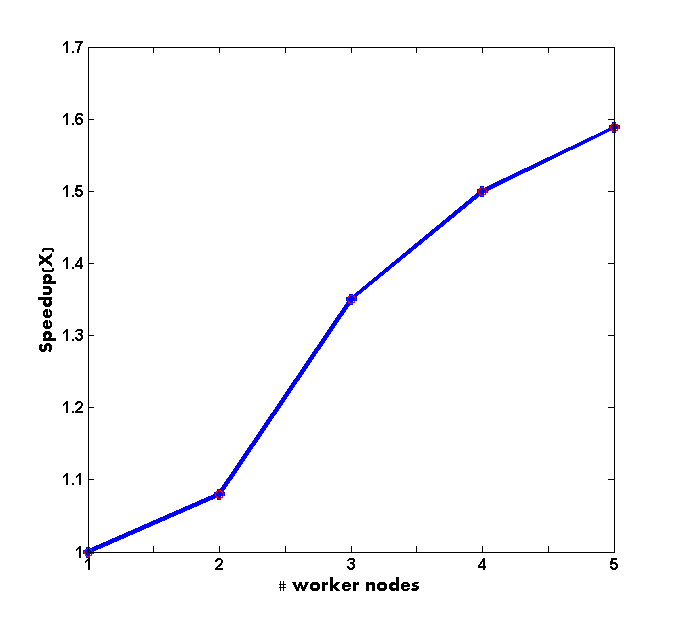}}
    \end{floatrow}%
  }
  
\end{figure}

We started by obtaining plots \ref{fig:f5} and \ref{fig:f6} which experimentally prove the convergence of our proposed algorithm. In Fig.\ref{fig:f5}, we see that the net error converges with the number of epochs while Fig.\ref{fig:f6} shows that the mean gradient values converge to zero with time as is expected. The results for cluster performance using FC Nets are plotted in Fig. \ref{fig:f36} (baseline runtime:14 mins(10k samples), 35 mins(0.1M samples)) and Fig. \ref{fig:h}; and using CNN are plotted in Fig. \ref{fig:f33} (baseline runtime:ranges from 11 mins(10k samples) to 75 hours(5M samples)) and Fig. \ref{fig:f34} for different sizes of the data samples. In all these plots, the number of model parameters is kept fixed and the number of data samples are varied.
In the second phase of experiments, we kept the number of data samples fixed and varied the number of model parameters for different network architectures. We have plotted the results for the variation of number of model parameters with number of partitions for CNN in Fig. \ref{fig:h8}. We were not able to accommodate more than 100 million parameters into the memory of any of the five single CPU machines. The partitions here correspond to the term "processes" which was used to explain the algorithm for model parallelism in \ref{sec} above.

We can see from the plot that as the size of model increases, we get more speedup. This is as expected because our algorithm performs well for large sizes of model parameters. Also, it can be seen that for a given model size, there is not much improvement in performance after a certain number of partitions.This is because we can't obtain speedup beyond a certain point in distributed systems. However, the saturation point increases as we increase the model sizes. For FC Nets and RNN/LSTM also, similar pattern of results was obtained, though the speedup obtained varied across networks. The speedup was more for CNN as compared to Fully Connected Nets because of reduced communication overhead between the machines due to the local connectivity structure in CNN.We also implemented a character-level predictor using RNN and LSTM. We have plotted the speedup obtained with increase in the number of worker machines in Fig. \ref{fig:h9} (baseline runtime: 8.1 hours). From the figure, it can be seen that the speedup increases uniformly with increase in the number of worker nodes thereby showing that our implementation of distributed LSTM is also scalable. The accuracy results of these implementations couldn't be included due to length constraints.

\section{Conclusions}
In this work, we have developed a generic distributed and scalable framework for training of deep networks in Apache Spark implementing both data parallelism and model parallelism. We have proposed a new algorithm for distributed backpropagation for model parallelism along with detailed cost analysis and mathematical proof of convergence. We have successfully applied our framework to different neural network architectures like Fully Connected Nets, CNN, RNN and LSTM. Through extensive simulations, we have concluded that our framework is pretty scalable, i.e, the performance improves as we add more nodes to the cluster. We have achieved about \textbf{11X speedup with 5M samples for CNN} (baseline: single machine), \textbf{7.2X speedup with 0.1M samples for Fully Connected Nets} and \textbf{5.6X speedup with 4 billion model parameters} just by using a cluster of five CPUs.We can conclude that our framework \textbf{performs best with increase in the size of data samples and large number of model parameters} (shown both mathematically and experimentally) and can be used conveniently for deep learning applications without requirement of expensive hardware. Hence, \textbf{our framework is distributed, scalable, economic and offers huge flexibility in terms of usage} due to its integration with Spark which is a generic batch computational framework. Its ability to rely on just a cluster of cheap commodity hardware widens the range of users further.

\bibliographystyle{splncs03}
\bibliography{biblio}

\section*{Appendix: Proof of Convergence}
Our goal is to minimize some parameter vector $x$ such that the sum over convex functions $f_i: X -> R$ takes on the smallest value possible. This can be written as Eq. (1) or (2) as given below:

\begin{equation}
\centering
    f^*(x) :=  1/|F| \sum_{i} f_i(x)
\end{equation}

\begin{equation}
\centering
   f^*(x) := E_{f~p(f)}[f(x)]
\end{equation}
 and correspondingly 

\begin{equation}
    x^* :=argmin_{x\in\chi} f^*(x)
\end{equation}     
We know that sum of convex functions in the same domain is also convex. These correspond to the below quantities in our distributed back propagation model:
\\Forward Pass $=> f_i$: sum of $a_i$'s from different machines at the master process with an  average delay $\tau_1$
\\Backward Pass $=> f_i$: sum of $\delta_j$'s from different machines at the master process with an average  delay $\tau_2$.

Since the calculation of gradient in back propagation consists of the forward and backward stages, we can write the net delay, $\tau$ as the weighted sum of delays during the two processes ($\alpha_1$ and $\alpha_2$ are scaling factor)

\begin{equation}
    \tau =\frac {\alpha_1\tau_1 + \alpha_2\tau_2} { \alpha_1 + \alpha_2}
\end{equation} 

In general the update rule in Stochastic Gradient Descent can be written as:

\begin{equation}
    w(t+1) = w(t) - \eta_t \Delta f_{t-\tau}(x)
\end{equation} 
 where, w(t) = weights at time t, $\eta_t$= Learning Rate, $\Delta f_t(x)$= Gradient of f(x) at time t. Bregman divergence is between $x$ and $x^*$ is defined as: 

\begin{equation}
    D(x||x^*) = 1/2 \parallel x-x^* \parallel^2
\end{equation} 

We directly take up three theorems defined in \cite{langford2009slow} to help us arrive at the final result.The detailed proof of these theorems can be found in \cite{langford2009slow}.

Theorem 1:  If we assume the cost function  f  to be Lipschitz continuous with a constant L and $max_{x,x^{'}\in\chi}$D($x \parallel x^{'}$) $\leq F^{2}$  given $\eta_{t} =\frac{\sigma}{\sqrt{t-\tau}}$ for some constant $\sigma >$ 0 and T to be the total number of iterations, the regret of the delayed update algorithm is bounded by

\begin{equation}
    R[X] \leq \sigma L^2\sqrt{T}  + F^2\frac{\sqrt{T}}{\sigma} + L^2 \frac{\sigma\tau^2}{2} + 2L^2\sigma\tau\sqrt{T}
\end{equation} 

Theorem 2: Suppose that the functions $f_i$ are strongly convex with parameter $\lambda > 0$. If we choose the learning rate as $\eta_t =\frac{\sigma}{\sqrt{t-\tau}}$  for t $ > \tau$  and $\eta_t = 0 $ for t $< \tau$, then under the assumptions of Theorem 1 we have the following bound:

\begin{equation}
    R[X] \leq \lambda \tau F^2 + [ \frac{1}{2} + \tau ]\frac{L^2}{\lambda}(1 + \tau + \log{T})"
\end{equation} 

A small delay should not significantly impact the update.This condition is equivalent to saying that a small change in the value of $x$ should not lead to major changes in values of the gradients.For this, we assume that the Lipschitz-continuity of the gradient of f.This can be stated as given below:

\begin{equation}
   \parallel \Delta f_t(x) - \Delta f_t(x^{'})\parallel \leq H \parallel x - x^{'}\parallel
\end{equation} 
Theorem 3: Under the assumptions of Theorem 2, in particular, assuming that all functions $f_i$ are i.i.d. and strongly convex with constant $\lambda$ and corresponding learning rate $\eta_t =\frac{\sigma}{\sqrt{t-\tau}}$ and Eq.(9) holds, we have the following bound on the expected regret:

\begin{equation}
 E[R[X]] \leq \frac{10}{9}[\lambda \tau F^2 + [ \frac{1}{2} \tau]\frac{L^2}{\lambda}[1 + \tau + \log{(3\tau +(H\tau/\lambda))}] +\frac{L^2}{2\lambda}[1 + \log{T}] + \frac{\pi^2 \tau^2 HL^2}{6\lambda^2}
\end{equation}

We have a \textbf{dependency of the form} $O(\tau\log{\tau}+\log{T})$. We get two separate terms dependent on $\tau$ and T. This implies that when T is large, delay $\tau$ essentially gets averaged out in different iterations and our algorithm converges with the upper bound as given above. \textbf{So, our algorithm will perform the best with more number of iterations (more data samples) and larger model sizes which is also borne out by the experimental results.}

\end{document}